\documentclass[a4paper,11pt]{article}
\usepackage{tikz}
\usepackage{ circuitikz }
\usetikzlibrary{decorations.pathreplacing}
\usepackage{pgfplots}
\usepackage{multirow,graphicx}

\usepackage[margin = 2.5cm]{geometry}
\usepackage{amssymb,amsthm,empheq,verbatim}
\usepackage{fancyhdr}
\usepackage{multirow}
\usepackage{epsfig}
\usepackage{graphicx,amsmath,float}
\usepackage[caption=false]{subfig}
\usepackage[colorlinks = false,
linkcolor = blue,
urlcolor  = blue,
citecolor = blue, 
anchorcolor=blue,
linkbordercolor=cyan,
urlbordercolor=cyan,
citebordercolor=cyan]{hyperref}
\usepackage{rotating, color,multicol}
\usepackage{fancyhdr,array,enumerate,cleveref}
\usepackage{enumitem}
\usepackage{todonotes}
\usepackage{eucal}
\usepackage{booktabs,stmaryrd}
\graphicspath{{figures/}}
\usepackage{setspace}
\usepackage{smartdiagram}
\usesmartdiagramlibrary{additions} 
\crefname{section}{Section}{Sections}
\crefname{figure}{Figure}{Figures}
\crefname{table}{Table}{Tables}
\crefformat{table}{#2Table~#1#3}
\crefformat{figure}{#2Figure~#1#3}
\crefformat{section}{#2Section~#1#3}

\usepackage[ruled,vlined]{algorithm2e}

\newcommand{\Id}{\operatorname{Id}}

\newcommand{\Fix}{\operatorname{Fix }}

\makeatletter
\newcommand{\ostar}{\mathbin{\mathpalette\make@circled\star}}
\newcommand{\make@circled}[2]{%
	\ooalign{$\m@th#1\smallbigcirc{#1}$\cr\hidewidth$\m@th#1#2$\hidewidth\cr}%
}
\newcommand{\smallbigcirc}[1]{%
	\vcenter{\hbox{\scalebox{0.77778}{$\m@th#1\bigcirc$}}}%
}

\definecolor{blue}{rgb}{0.0, 0.0, 0.55}
\definecolor{beamer@blendedblue}{rgb}{0.2,0.5,0.6}
\definecolor{mygreen}{rgb}{0.2431, 0.5882, 0.3176}
\definecolor{mypurple} {rgb}{0.4196, 0.2980, 0.6039}

\newcommand{\keywords}[1]{\textbf{Keywords:} #1}

\begin{document}

	\title{Holistic Processing of Colour Images Using Novel Quaternion-Valued Wavelets on the Plane}	
	\author{Neil D. Dizon\\University of Helsinki \and Jeffrey A. Hogan\\University of Newcastle}
	
	\date{January 11, 2024}
	
	\maketitle	
	
\begin{abstract}Recently, novel quaternion-valued wavelets on the plane were constructed using an optimisation approach. These wavelets are compactly supported, smooth, orthonormal, non-separable and truly quaternionic. However, they have not been tested in application. In this paper, we introduce a methodology for decomposing and reconstructing colour images using quaternionic wavelet filters associated to recently developed quaternion-valued wavelets on the plane. We investigate its applicability in compression, enhancement, segmentation, and denoising of colour images. Our results demonstrate these wavelets as promising tools for an end-to-end quaternion processing of colour images.
\end{abstract}


\thispagestyle{fancy}
\section{Introduction}

Wavelets have long been known as a powerful tool for analysing and processing greyscale images. With their ability to decompose an image into different scales, one can extract important information from an image that can then be used in a variety of applications including compression, denoising, enhancement, feature extraction, registration, and segmentation. By treating each channel of a multi-channel image as greyscale, wavelet-based image processing schemes have also been extended to multi-channel signals like colour images. 

The most basic model of a colour image is a three-channel image consisting of red, green, and blue (RGB) components of the pixels. Other commonly used models include the luminance-chrominance (YUV) and cyan-magenta-yellow-key (CMYK) that use three and four channels, respectively. Other four-channel signal models like RGB-A and RGB-NIR are also becoming more prevalent. In these models, the fourth band corresponds to an \emph{alpha} and a \emph{near-infrared} (NIR) component, respectively. Most of the present-day handling of colour images relies on the analysis of each channel separately. With this kind of approach, the possible correlations between the channels are totally ignored, if not undervalued. It is preferable to encode the pixel components into higher-dimensional algebras which are anticipated to exploit correlation between channels. For the case of colour images with three or four channels, the algebra of \emph{quaternions} is sufficient \cite{peicolour,sangwinecolour}. But once a higher-dimensional signal is embedded into this algebra, more sophisticated wavelet transforms become imperative.

In the literature, a number of published articles considered different extensions of wavelet transforms to the quaternionic setting. Fletcher and Sangwine \cite{fletchersangwine} noted in their survey on the development of  quaternionic wavelet transforms (QWT) that most of these extensions were just derived from real filter coefficients, and are just separate discrete or complex wavelet transforms in disguise. They further noted that the extensions due to Hogan and Morris \cite{hoganmorris1}, and Ginzberg and Walden \cite{ginzbergwalden} are among the few that attempted to develop true QWT. Recently, Fletcher \cite{fletcher} extended Ginzberg's work to construct examples of quaternion-valued scaling filters on the line.  

The quaternionic wavelet theory developed by Hogan and Morris \cite{hoganmorris1,morristhesis} provided direct analogues of classical wavelet theory for construction of quaternion-valued wavelets on the plane. The quaternionic quadrature mirror filter conditions (QQMF) and the scaling equation for quaternionic wavelets were rephrased through the notion of spinor-vector matrices. They have also derived quaternionic counterparts of compact support, orthonormality, and regularity conditions. However, no examples of quaternionic wavelets satisfying these properties were constructed.

In a different pursuit, Franklin, Hogan, and Tam \cite{FHTconference,FHTpaper,franklinthesis} developed techniques that have been successful in reproducing Daubechies’ wavelets using an optimisation approach. In particular, wavelet architecture was formulated as a \emph{feasibility problem} of finding a point on the intersection of constraint sets arising from the design criteria and the conditions of multiresolution analysis (MRA). This feasibility approach to wavelet construction has successfully produced new examples of non-separable, complex-valued, smooth, compactly supported, orthonormal wavelets on the plane.

Inspired by the extendability of the feasibility approach to higher-dimensional constructions, Dizon and Hogan \cite{dizonthesis,DizonHoganQuat} revisited the quaternionic wavelet theory developed by Hogan and Morris. They formulated  and solved the construction of quaternionic wavelets as feasibility problems.  Solutions to these feasibility problems admit novel examples of quaternion-valued wavelets on the plane (refer to \cref{fig:quat_wavelets} for an example). The successful architecture of compactly supported, smooth and orthonormal quaternion-valued wavelets on the plane leaves open many important avenues of research. With these wavelets, the pixel components of a colour image may now be encoded into the scalar and imaginary parts of quaternions for holistic processing of signals using wavelet transforms. We use the term \emph{holistic} to mean that the components of a pixel from different channels are treated as a whole rather than separately \cite{fletchersangwine,alfsmann}. With such an approach, the potentially useful correlations between the pixel components are not lost. 

However, the work done in \cite{dizonthesis, DizonHoganQuat} focused solely on constructing quaternion-valued wavelets on the plane and did not delve into application. The development and formalisation of a suitable quaternion-valued wavelet decomposition and reconstruction is still left lacking. In this regard, our current work distinguishes itself from the aforementioned papers as we now aim to look into the applicability of these wavelets.

The contributions of this paper are summarised as follows: (i) we develop a scheme that decomposes and reconstructs colour images using quaternionic scaling and wavelet filters associated to the recently developed quaternion-valued wavelets on the plane; and (ii) we exemplify some image processing steps that can be done in between wavelet decomposition and reconstruction to allow for compression, enhancement, segmentation, and denoising of colour images. In the context of applications, our primary objective is to elucidate the potential of employing a holistic image processing methodology using these novel quaternion-valued wavelets. It is important to note that our intention is to emphasise the inherent promise of this approach, rather than to ascertain any superiority in performance, which we reserve for future work.

The rest of the paper is organised as follows. In \cref{sec:quaternionwavelets}, we revisit the feasibility approach for the construction of quaternion-valued wavelets on the plane with the goal of highlighting their important properties. \cref{sec:decompositionreconstruction} formalises the decomposition and reconstruction scheme for colour images using quaternionic scaling and wavelet filters. Here, we illustrate energy compaction in the decomposition, and demonstrate perfect reconstruction when no alterations were made in the wavelet coefficients. In \cref{sec:applications}, we present image processing applications in line with the stated contribution of this paper.

\section{Quaternion-valued wavelets on the plane}
\label{sec:quaternionwavelets}

Recently, Dizon and Hogan constructed quaternion-valued wavelets on the plane through the feasibility approach (for a detailed discussion, see \cite{dizonthesis,DizonHoganQuat}). The construction entails formulating wavelet architecture as feasibility problems.

A \emph{feasibility problem} is a special type of optimisation problem that seeks to find a point in the intersection of a finite family of sets. Formally, given sets $K_1, K_2, \dots, K_r$ contained in a Hilbert space $\mathcal{H}$, the corresponding feasibility problem is defined by:
\begin{equation*}
	\text{find~} x^\ast \in K :=\bigcap_{j=1}^r K_j. 
\end{equation*}
In the literature, the method of alternating projections (MAP) \cite{neumann} and the {Douglas--Rachford} (DR) algorithm \cite{drachford} are well-known examples of \emph{projection algorithms} that are able to solve two-set feasibility problems.  Both algorithms are amenable to solve many-set feasibility problems through Pierra's {product space reformulation} \cite{pierra}. 

\begin{figure}[h!]
    \centering
			\begin{tikzpicture}[x=0.75cm,y=0.75cm, scale=1]
				\begin{tiny}
					\clip(-5.71644693762893,-3.6027777371602863) rectangle (5.562396356753598,4.546678084553665);
					\draw [line width=0.015pt,color=mypurple,fill=mypurple,fill opacity=0.3] (2.,0.) circle (2.25cm);
					\draw [line width=0.015pt,color=green,fill=green,fill opacity=0.3] (-2.,0.) circle (1.95cm);
					\node [] at (2,-2.5) {$K_2$}; 
					\node [] at (-2,-2) {$K_1$}; 
					
					\draw [fill=black] (-5.234682834894676,3.9968634686346913) circle (1.5pt);
					\draw[color=black] (-5.134146134013121,4.206591747735668) node {$x_0$};
					
					\draw [black,->] (-5.234682834894676,3.9968634686346913) -- (-2.0366143811507778,0.045241740819073506);

					\draw [fill=black] (-2.0366143811507778,0.045241740819073506) circle (1.5pt);
					\draw[color=black] (-2.8,-0.3) node {$R_{K_1}(x_0)$};
					
					\draw [black,->] (-2.0366143811507778,0.045241740819073506) -- (0.0369911928948925,0.022001094801453278);
					
					\draw [fill=black] (0.0369911928948925,0.022001094801453278) circle (1.5pt);
					\draw[color=black] (1.1,0.5) node {$R_{K_2}(R_{K_1}(x_0))$};

					\draw [line width=0.4pt] (-5.234682834894676,3.9968634686346913)-- (0.0369911928948925,0.022001094801453278);
					
					\draw [fill=black] (-2.598845820999892,2.0094322817180723) circle (1.5pt);
					\draw[color=black] (-2.3,2.4) node {$x_1$};
					
				\end{tiny}
			\end{tikzpicture}
            \caption{One step of a Douglas--Rachford fixed-point iteration which follows a simple \emph{reflect-reflect-average} scheme. Starting with a point $x_0$, the algorithm performs a reflection with respect to $K_1$ to obtain the point $R_{K_1}(x_0)$, followed by another reflection with respect to $K_2$ to obtain the point $R_{K_2}(R_{K_1}(x_0))$. Averaging $x_0$ and $R_{K_2}(R_{K_1}(x_0))$ yields the point corresponding to the next iterate $x_1$. }
            \label{fig:DRstep}
\end{figure}
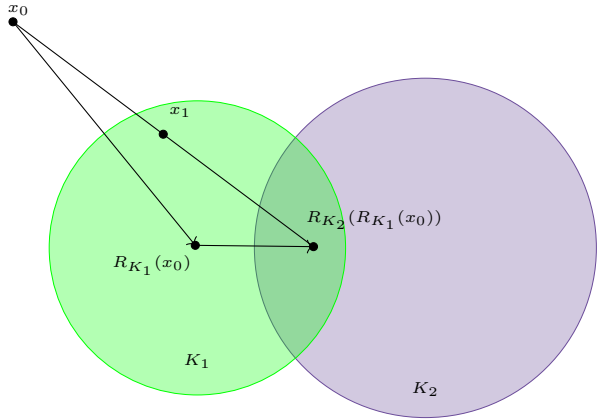

The Douglas--Rachford method has been observed to exhibit empirical potency even in nonconvex settings \cite{aacampoy,bsims,dtam}. Like most projection algorithms, DR exploits the concept of projectors and reflectors. If $C$ is a nonempty subset of $\mathcal{H}$, the \emph{projector} onto $C$ is the set-valued operator $P_{C}\colon \mathcal{H} \rightrightarrows C$ defined by
		\begin{equation*}
			P_{C}(x) = \{ c \in C: \|x-c\| = \inf_{z\in C}\|x-z\| \};
		\end{equation*}
		and the \emph{reflector}  with respect to $C$ is the set-valued operator $R_{C}\colon \mathcal{H} \rightrightarrows \mathcal{H}$ defined by
		\begin{equation*}
			R_{C} := 2P_{C} - \Id,
		\end{equation*}
		where $\Id$ denotes the identity map. An element of $P_{C}(x)$ is called a \emph{projection} of $x$ onto $C$. Similarly, an element of $R_C(x)$ is called a \emph{reflection} of $x$ with respect to $C$. Note that use of ``$\rightrightarrows$'' is to emphasise that an operator is (possibly) set-valued.  Formally, given two nonempty subsets $K_1$ and $K_2$ of $\mathcal{H}$, the \emph{DR operator} $T_{K_1,K_2}$ is defined as
	\begin{equation*}
		T_{K_1,K_2} := \frac{\Id+R_{K_2}R_{K_1}}{2}.
	\end{equation*}
If $K_1$ and $K_2$ are closed convex subsets of $\mathcal{H}$ with $K_1\cap K_2 \neq \varnothing$, then for any $x_0 \in \mathcal{H}$, the sequence $(x_n)_{n\in \mathbb{N}}$ generated by $x_{n+1} = T_{K_1,K_2}(x_n)$ converges weakly to a point $x^\ast \in \Fix T_{K_1,K_2}$, and the \emph{shadow sequence} $(P_{K_1}(x_n))_{n\in \mathbb{N}}$ converges weakly to $P_{K_1}(x^\ast) \in K_1\cap K_2$ \cite{lions,svaiter}. Refer to \cref{fig:DRstep} for a simple illustration of the Douglas--Rachford scheme on two sets. 

In wavelet feasibility problems, the constraint sets encode the basic \emph{compact support}, \emph{orthonormality}, and \emph{regularity} conditions. The feasibility approach to wavelet construction treats these design criteria as constraints that must be simultaneously satisfied. Such a technique has been also successful in reproducing Daubechies’ wavelets, and in deriving \emph{non-separable} examples of complex-valued, compactly supported, smooth and orthonormal wavelets on the plane \cite{FHTconference,FHTpaper,franklinthesis}. We note here that the compact support of the scaling and wavelet functions facilitate speedy and accurate computation of transform coefficients in the wavelet decomposition of a given image signal. In applications, it is also preferred that wavelets have continuous and bounded derivatives as this property allows for more parsimonious expansions.  Additional constraints can be imposed to promote \emph{symmetry} which helps alleviate distortion around edges in images \cite{daubechiesbook,daubechies1}. Moreover, the term non-separable means that these higher-dimensional wavelets are not formed as tensor products of wavelets in a lower-dimensional space. Non-separable wavelets are preferred since they have isotropic characteristics that avoid partiality toward the coordinate directions \cite{cohendaubechies,kovacevic,laisurvey}.

The feasibility problem formulation becomes even more challenging and intricate for quaternion-valued wavelets on the plane, primarily because of the increased dimensionality and with the absence of commutativity as an additional complicating factor. For a comprehensive discussion on the quaternionic wavelet feasibility problem, refer to \cite{dizonthesis,DizonHoganQuat}.

\begin{figure}[h!]
	\centering
	\includegraphics[width=0.8\linewidth]{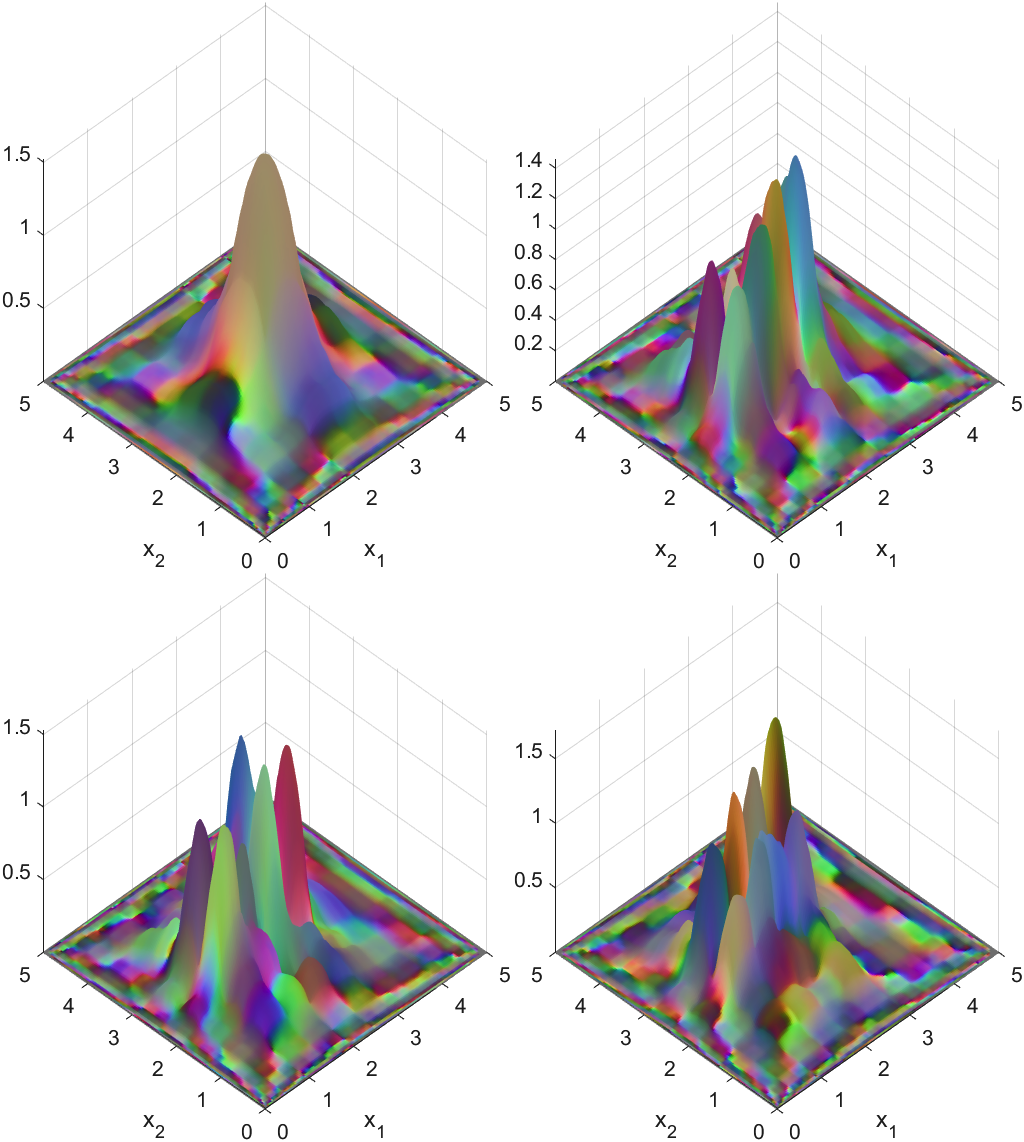}
	\caption{An example of a wavelet ensemble generated from a solution of the quaternionic wavelet feasibility problem. For each plot, the height of a point on the graph corresponds to the modulus of the quaternion, and the intensities of RGB colour of the point represent the imaginary parts in the polar form of the quaternion.}
	\label{fig:quat_wavelets}
\end{figure}

\cref{fig:quat_wavelets} shows an example of a quaternion-valued wavelet ensemble on the plane derived as a solution to the quaternionic wavelet feasibility problem. A \emph{wavelet ensemble} consists of a scaling function and three associated wavelets. These functions are compactly supported, smooth, orthonormal, non-separable, and truly quaternionic. Additionally, the scaling function is pointwise symmetric about its centre of support. These functions are further associated to their respective filters, i.e., the \emph{scaling and wavelet filters}\footnote{Relevant source codes, solutions, and wavelet filters are available at \url{https://gitlab.com/nddizon1/waveletconstruction}, where we have also used the Quaternion Toolbox for MATLAB\textsuperscript{\textregistered} \cite{qtfm}.}. Throughout this paper, we only use the scaling and wavelet filters associated to the wavelet ensemble in \cref{fig:quat_wavelets}. Other wavelet ensembles are presented in \cite[Chapter~8]{dizonthesis}, derived as solutions to quaternionic wavelet feasibility problems.

To understand how quaternion-valued wavelets on the plane were plotted \cref{fig:quat_wavelets}, we first define the set $\mathbb{H}$ of quaternions by 
\[
\mathbb{H} := \big\{a+be_1+ce_2+de_{12} \,:\, a,b,c,d \in \mathbb{R}, \, e_{1}^2 = e_2^2=e_{12}^2=e_1e_2e_{12}=-1\big\}
\]
 where we use $e_1, e_2$ and $e_{12}$ to denote the imaginary units. In plotting quaternion-valued wavelets (or any quaternion-valued functions) on the plane, we used the following idea. For any set $X \subseteq \mathbb{R}^2$, let $f: X \to \mathbb{H}$ be a quaternion-valued function, i.e., $f(x) = f_0(x) + f_1(x)e_1 + f_2(x)e_{2} + f_{12}(x)e_{12}$ where $f_0,f_1,f_2,f_{12}: X \to \mathbb{R}$. For a fixed $x=(x_1,x_2) \in X$, we write $f(x) = |f(x)|e^{\mu_{f(x)} \phi_{f(x)}}$ in polar form. Since $\mu_{f(x)} \phi_{f(x)}$ is a pure quaternion (i.e., its real part is zero), we can write it as $$\mu_{f(x)} \phi_{f(x)} = R_{f(x)}e_1 + G_{f(x)}e_2 + B_{f(x)}e_{12}$$ with $R_{f(x)},G_{f(x)},B_{f(x)}$ the corresponding imaginary parts of $\mu_{f(x)} \phi_{f(x)}$. Thus, we may associate $(x,f(x))$ with a point in $\mathbb{R}^3$ with coordinates $(x_1,x_2, |f(x)|)$ and coloured by $(R_{f(x)},G_{f(x)},B_{f(x)})$ injected into the RGB colour space.

\section{Decomposition and reconstruction using quaternionic filters}
\label{sec:decompositionreconstruction}

Colour image processing with quaternion-valued wavelets relies on a suitable wavelet decomposition and reconstruction using scaling and wavelet filters. In between the decomposition and reconstruction steps, several image processing tasks may be implemented including (but not limited to) compression, enhancement, segmentation, and denoising.

In this section, we formalise how colour images can be embedded into the algebra of quaternions.  We start with the RGB colour image model but eventually add a near-infrared (NIR) channel for consideration of RGB-NIR images. After this, we describe a suitable decomposition and reconstruction scheme using quaternionic scaling and wavelet filters.

\subsection{Colour images and quaternion algebra}
\label{subsec:colourimages}

Typically, an RGB colour image is viewed as a function $F: \mathbb{R}^2 \to \mathbb{R}^3$ given by 
$F(x) = ({R}(x), {G}(x), {B}(x))$
where $R(x)$, $G(x)$ and $B(x)$ are the {red}, {green} and {blue} components of the pixel $x$, respectively.

Alternatively, with the algebra of quaternions $\mathbb{H}$, we may view an RGB colour image as a quaternion-valued function $F: \mathbb{R}^2 \to \mathbb{H}$ given by
$$F(x)={R}(x)e_1 + {G}(x)e_2 + {B}(x)e_{12}$$
where the red, green and blue components are embedded into the imaginary parts of a quaternion. This concept of representing colour image pixels using pure quaternions has been independently developed on two occasions: by Sangwine \cite{sangwine1996fourier}, and by Pei and Cheng \cite{pei1997novel}, for developing colour image compression algorithms. Such a representation simultaneously handles and distinguishes between the internal multivariate nature of the RGB colour image (i.e., each pixel being a 3D vector) and its external multidimensional nature (i.e., the whole image being a multidimensional array of spatial pixels) in an elegant way. On the other hand, the corresponding representation in the real domain is frequently unwieldy and typically managed by stacking the three or four components as vectors or matrices \cite{miron2023quat}. However, doing so means completely ignoring the potential relationships between the internal components and the geometric characteristics of the vector data. Conversely, the algebraic encoding of 3D vectors using quaternions allows for the innate representation of vector data with multiple dimensions as quaternion vectors or matrices.

If we are given a 4D vector signal, then we can make use of the full quaternion representation. An example of such a 4D vector signal is an RGB-NIR image. Embedding the near-infrared component into the real part of a quaternion, an RGB-NIR image can be viewed as a full-quaternion-valued function  $F: \mathbb{R}^2 \to \mathbb{H}$ given by
$$F(x)=I(x) + {R}(x)e_1 + {G}(x)e_2 + {B}(x)e_{12},$$
where $I(x)$ represents the near-infrared component of the pixel $x$. Throughout this paper, we use RGB-NIR images as our primary example. Note that other colour image models with four channels can be considered in a similar manner.

\subsection{Quaternionic wavelet decomposition and reconstruction}

In order to be able to use quaternion-valued wavelets on the plane to process RGB-NIR images, we first formalise a suitable decomposition and reconstruction scheme. Henceforth, we denote the scaling filter by $H$, and the three wavelet filters by $G_1, G_2$ and $G_3$. Below are the proposed steps in processing RGB-NIR images using quaternion-valued wavelets.

\begin{enumerate}
    \item \textit{Quaternion embedding:} In this process, an RGB-NIR image is transformed into a matrix with quaternion entries as described in \cref{subsec:colourimages}. We refer to this matrix as the quaternion representation of our colour image.
    
    \item \textit{Decomposition:} This is carried out by convolving the quaternion representation with one low-pass (scaling) and three high-pass (wavelet) filters, followed by downsampling. The resulting coefficients from the low-pass filtering contains the low-frequency content or approximation of the original colour image, while the other three coefficients capture the high-frequency details. We repeat the filtering and downsampling process on the approximation coefficients until we achieve the desired depth of decomposition.
    
    \item \textit{Image processing: } This involves modifying the wavelet coefficients in various ways, with each alteration tailored to the specific image processing objective at hand. Detailed descriptions of basic image processing tasks considered in this paper are provided in \cref{sec:applications}.

    \item \textit{Reconstruction: } From the altered wavelet coefficients,  we perform upsampling and inverse filtering to each set of coefficients at each level. 
    
    \item \text{\textit{Channel extraction}:} Since the previous step yields a matrix with quaternion entries, we extract the colour image channels of the processed image from the real and imaginary parts of the quaternionic matrix.
\end{enumerate}

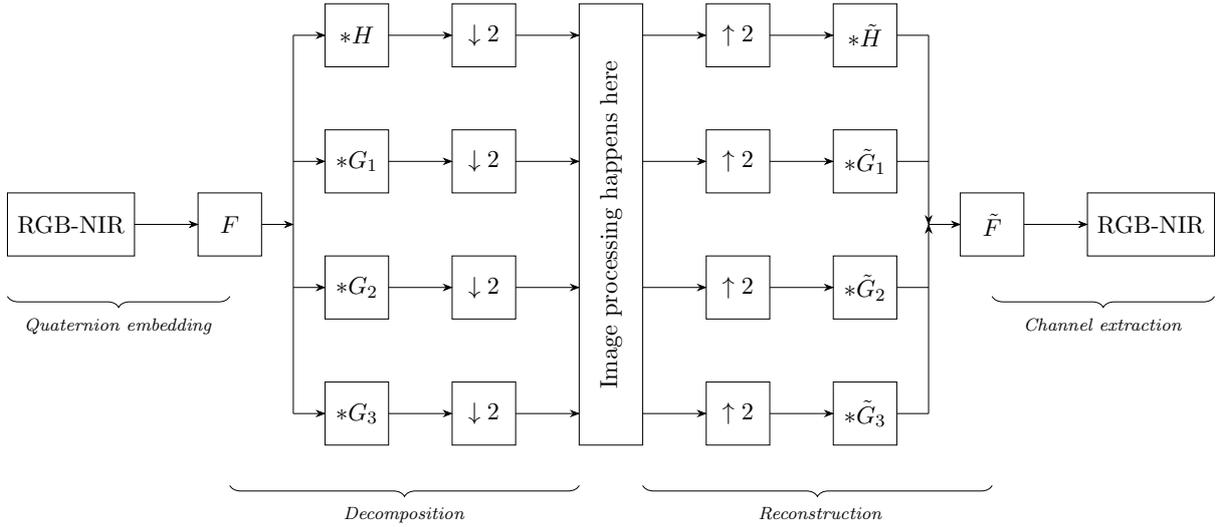
\begin{figure}[!h]
%
%
\centering
\resizebox{\textwidth}{!}{%
\begin{tikzpicture}
\tikzstyle{every node}=[font=\normalsize]

\draw [, line width=0.2pt ] (-3,0) rectangle  node {\normalsize RGB-NIR} (-1,1);
\draw [ line width=0.2pt, -Stealth] (-1.0,0.5) -- (0.0,0.5);
\draw [, line width=0.2pt ] (0,0) rectangle  node {\normalsize $F$} (1,1);

\draw [ line width=0.2pt, -Stealth] (1.0,0.5) -- (1.5,0.5);
\draw [ line width=0.2pt] (1.5,0.5) -- (1.5,-2.5);
\draw [ line width=0.2pt] (1.5,0.5) -- (1.5,3.5);
\draw [ line width=0.2pt, -Stealth] (1.5,3.5) -- (2,3.5);
\draw [ line width=0.2pt, -Stealth] (1.5,1.5) -- (2,1.5);
\draw [ line width=0.2pt, -Stealth] (1.5,-0.5) -- (2,-0.5);
\draw [ line width=0.2pt, -Stealth] (1.5,-2.5) -- (2,-2.5);

\draw [, line width=0.2pt ] (2,3) rectangle  node {\normalsize $*H$} (3,4);
\draw [, line width=0.2pt ] (2,1) rectangle  node {\normalsize $*G_1$} (3,2);
\draw [, line width=0.2pt ] (2,-1) rectangle  node {\normalsize $*G_2$} (3,0);
\draw [, line width=0.2pt ] (2,-3) rectangle  node {\normalsize $*G_3$} (3,-2);

\draw [ line width=0.2pt, -Stealth] (3,3.5) -- (4,3.5);
\draw [ line width=0.2pt, -Stealth] (3,1.5) -- (4,1.5);
\draw [ line width=0.2pt, -Stealth] (3,-0.5) -- (4,-0.5);
\draw [ line width=0.2pt, -Stealth] (3,-2.5) -- (4,-2.5);

\draw [, line width=0.2pt ] (4,3) rectangle  node {\normalsize $\downarrow 2$} (5,4);
\draw [, line width=0.2pt ] (4,1) rectangle  node {\normalsize $\downarrow 2$} (5,2);
\draw [, line width=0.2pt ] (4,-1) rectangle  node {\normalsize $\downarrow 2$} (5,0);
\draw [, line width=0.2pt ] (4,-3) rectangle  node {\normalsize $\downarrow 2$} (5,-2);

\draw [ line width=0.2pt, -Stealth] (5,3.5) -- (6,3.5);
\draw [ line width=0.2pt, -Stealth] (5,1.5) -- (6,1.5);
\draw [ line width=0.2pt, -Stealth] (5,-0.5) -- (6,-0.5);
\draw [ line width=0.2pt, -Stealth] (5,-2.5) -- (6,-2.5);

\draw [, line width=0.2pt ] (6,-3) rectangle  node [rotate=90] {\normalsize Image processing happens here} (7,4);

\draw [ line width=0.2pt, -Stealth] (7,3.5) -- (8,3.5);
\draw [ line width=0.2pt, -Stealth] (7,1.5) -- (8,1.5);
\draw [ line width=0.2pt, -Stealth] (7,-0.5) -- (8,-0.5);
\draw [ line width=0.2pt, -Stealth] (7,-2.5) -- (8,-2.5);

\draw [, line width=0.2pt ] (8,3) rectangle  node {\normalsize $\uparrow 2$} (9,4);
\draw [, line width=0.2pt ] (8,1) rectangle  node {\normalsize $\uparrow 2$} (9,2);
\draw [, line width=0.2pt ] (8,-1) rectangle  node {\normalsize $\uparrow 2$} (9,0);
\draw [, line width=0.2pt ] (8,-3) rectangle  node {\normalsize $\uparrow 2$} (9,-2);

\draw [ line width=0.2pt, -Stealth] (9,3.5) -- (10,3.5);
\draw [ line width=0.2pt, -Stealth] (9,1.5) -- (10,1.5);
\draw [ line width=0.2pt, -Stealth] (9,-0.5) -- (10,-0.5);
\draw [ line width=0.2pt, -Stealth] (9,-2.5) -- (10,-2.5);

\draw [, line width=0.2pt ] (10,3) rectangle  node {\normalsize $*\tilde H$} (11,4);
\draw [, line width=0.2pt ] (10,1) rectangle  node {\normalsize $*\tilde G_1$} (11,2);
\draw [, line width=0.2pt ] (10,-1) rectangle  node {\normalsize $*\tilde G_2$} (11,0);
\draw [, line width=0.2pt ] (10,-3) rectangle  node {\normalsize $* \tilde G_3$} (11,-2);

\draw [ line width=0.2pt] (11,3.5) -- (11.5,3.5);
\draw [ line width=0.2pt] (11,1.5) -- (11.5,1.5);
\draw [ line width=0.2pt] (11,-0.5) -- (11.5,-0.5);
\draw [ line width=0.2pt] (11,-2.5) -- (11.5,-2.5);

\draw [ line width=0.2pt, -Stealth] (11.5,3.5) -- (11.5,0.5);
\draw [ line width=0.2pt, -Stealth] (11.5,-2.5) -- (11.5,0.5);
\draw [ line width=0.2pt, -Stealth] (11.5,0.5) -- (12,0.5);

\draw [, line width=0.2pt ] (12,0) rectangle  node {\normalsize $\tilde F$} (13,1);

\draw [ line width=0.2pt, -Stealth] (13.0,0.5) -- (14.0,0.5);
\draw [, line width=0.2pt ] (14,0) rectangle  node {\normalsize RGB-NIR} (16,1);

\draw [decorate,decoration={brace,amplitude=5pt,mirror,raise=4pt}] (-3.0,-0.5) -- (0.5,-0.5) node [midway,below,yshift=-10pt] {\scriptsize \textit{Quaternion embedding}};

\draw [decorate,decoration={brace,amplitude=5pt,mirror,raise=4pt}] (0.5,-3.5) -- (6,-3.5) node [midway,below,yshift=-10pt] {\scriptsize \textit{Decomposition}};

\draw [decorate,decoration={brace,amplitude=5pt,mirror,raise=4pt}] (7,-3.5) -- (12.5,-3.5) node [midway,below,yshift=-10pt] {\scriptsize\textit{ Reconstruction}};
\draw [decorate,decoration={brace,amplitude=5pt,mirror,raise=4pt}] (12.5,-0.5) -- (16.0,-0.5) node [midway,below,yshift=-10pt] {\scriptsize \textit{Channel extraction}};
\end{tikzpicture}

}%
\caption{Colour image processing steps using quaternion-valued wavelets on the plane. The basic layout of a one-level discrete wavelet transform (with scaling filter $H$, wavelet filters $G_1, G_2, G_3$, and their respective inverse filters $\tilde{H}, \tilde{G}_1, \tilde{G}_2, \tilde{G}_3$) also includes the downsampling  ($\downarrow 2$) and upsampling ($\uparrow 2$) steps.}
\label{fig:QDWT}
\end{figure}

For a schematic diagram of a one-level decomposition and reconstruction, refer to \cref{fig:QDWT}. It is worth nothing that we simply followed Mallat's algorithm to carry out the decomposition and reconstruction \cite{mallat}. The simplicity of this wavelet decomposition and reconstruction scheme is a consequence of the fact that the new quaternion-valued wavelets that we are using are direct analogues of Daubechies' wavelets in the quaternion setting. Additionally, we note that the scaling function and wavelets (and hence, the scaling and wavelet filters) are not tensor products of lower-dimensional wavelets (wavelet filters), and that their filter coefficients or weights are quaternionic.

\begin{figure}[H]
	\centering
    \includegraphics[width=0.8\linewidth]{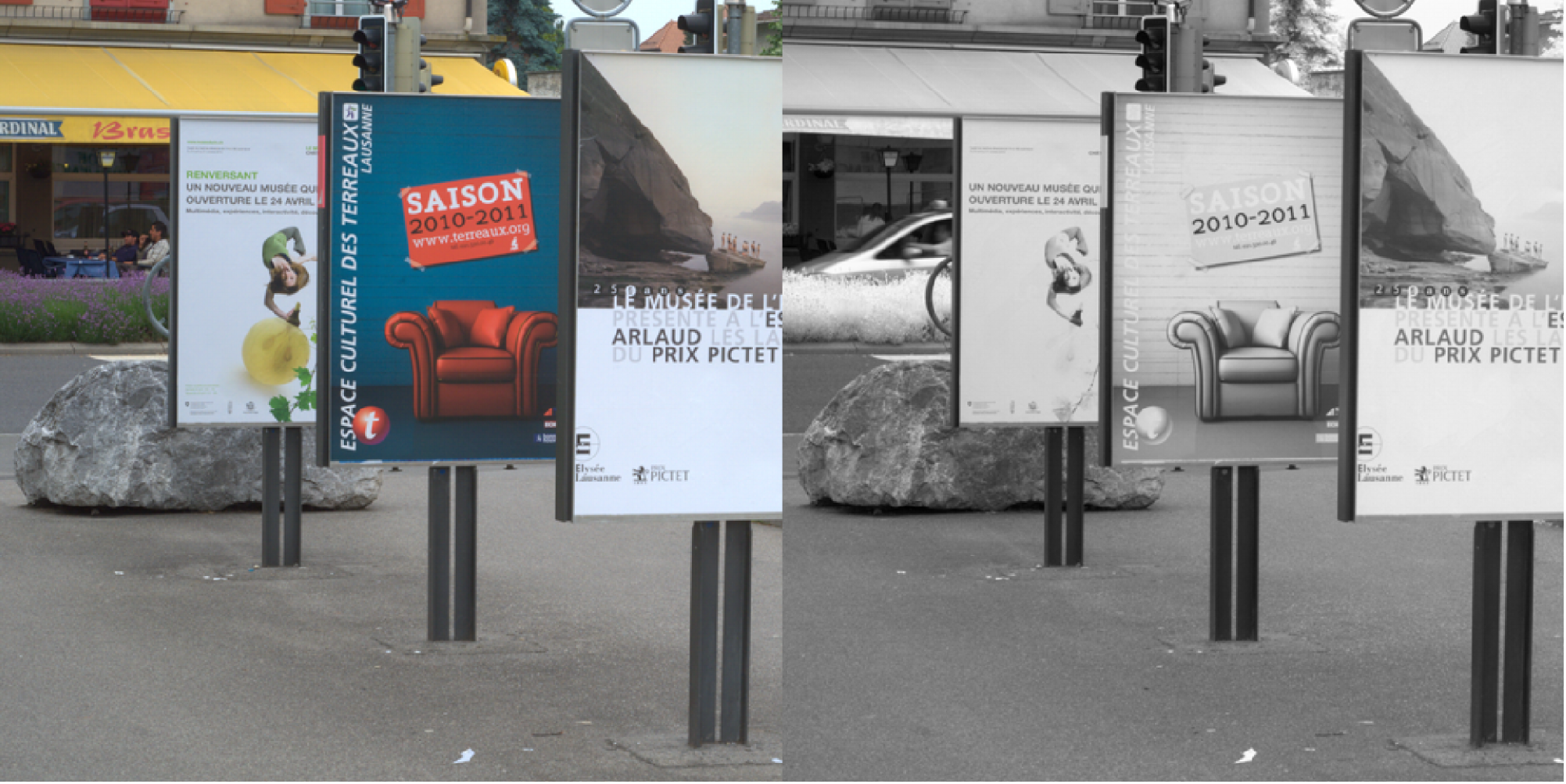}\hfill
	\caption{Example RGB-NIR image. This also coincides with the reconstructed image when no alternations were made in the wavelet coefficients.}
	\label{fig:reconstructed}
\end{figure}

Throughout this paper, we consider as an example\footnote{Resized image is borrowed from the \emph{RGB-NIR Scene Dataset} which is publicly available at \url{https://ivrlwww.epfl.ch/supplementary_material/cvpr11/index.html}.} the given $512 \times 512$ RGB-NIR image in \cref{fig:reconstructed}. All experiments were performed using MATLAB\textsuperscript{\textregistered} on a mid-2022 MacBook Air, equipped with 16GB of RAM and 8 CPU cores. In reporting image quality, we use built-in  MATLAB commands for full-reference algorithms including peak signal-to-noise ratio (PSNR) and structural similarity index (SSIM). We also use no-reference quality algorithms\footnote{Since these algorithms accept an RGB or a greyscale image, we report the average for RGB and NIR compenents.} like blind or referenceless image spatial quality evaluator (BRISQUE), natural image quality evaluator (NIQE) and perception based image quality evaluator (PIQE).

\begin{figure}[h!]
	\centering
    \includegraphics[width=0.8\linewidth]{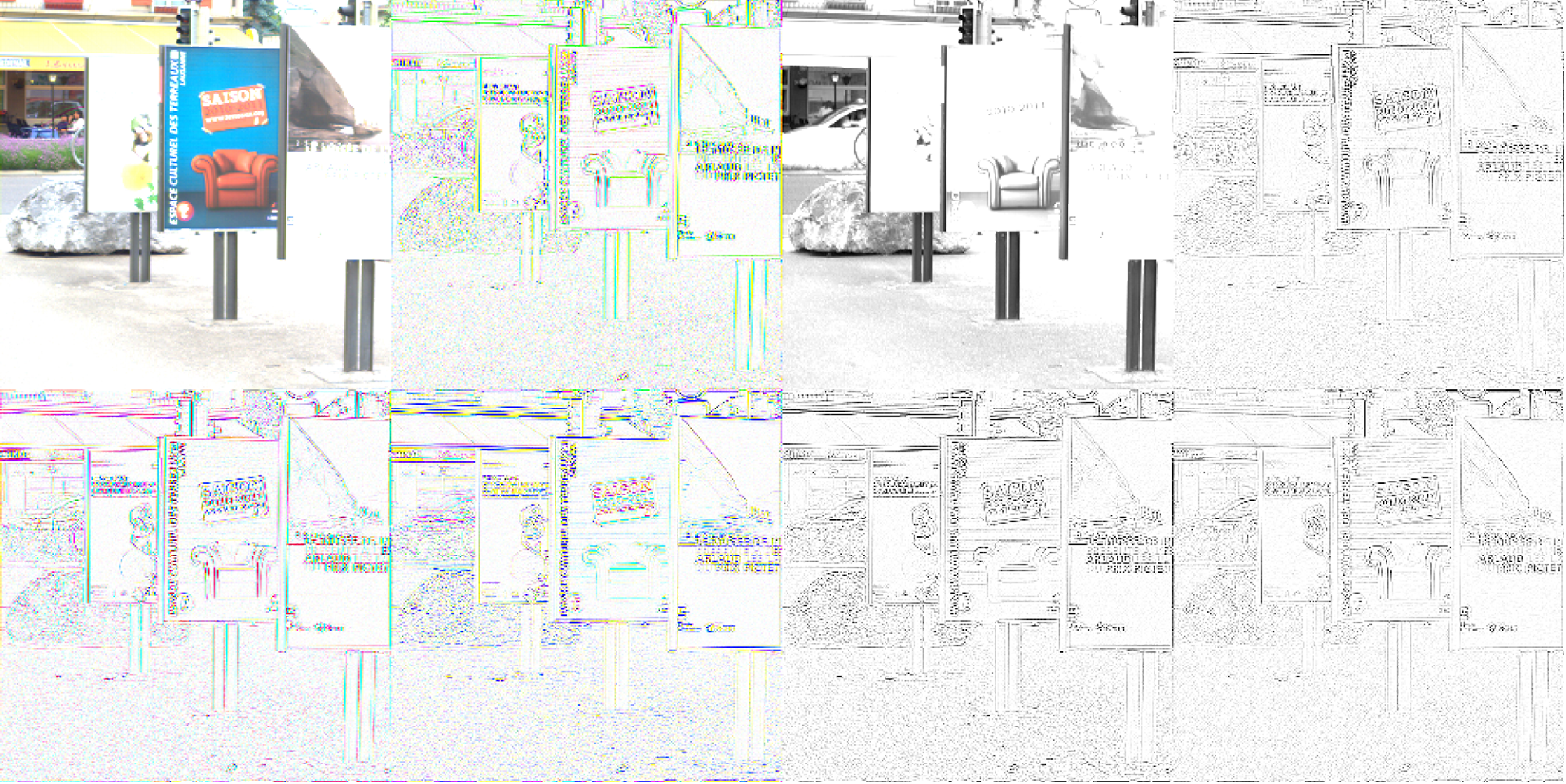}\hfill
	\caption{Example of a one-level wavelet decomposition of an RGB-NIR image (with detail coefficient values inverted and accentuated for illustrative purposes). The colour image on the left displays the imaginary parts of the quaternionic wavelet decomposition treated as RGB, while the greyscale image on the right is the scalar part of the decomposition.}
	\label{fig:decomposed}
\end{figure}

In \cref{fig:decomposed}, we present a one-level wavelet decomposition of our example RGB-NIR image. When no image processing is done in between the decomposition and reconstruction, the reconstructed image will perfectly coincide with the originally given colour image. The image quality scores of the reconstructed (and hence the original) RGB-NIR image are detailed in \cref{tab:image_scores}.

Finally, we also highlight that when using non-separable quaternion-valued wavelets, the energy in the wavelet decomposition is distributed fairly among the detail coefficients. This equitable distribution suggests that these quaternion-valued wavelets do not show preferential treatment toward any specific coordinate directions.

\subsection{Energy compaction in the wavelet domain}

Similar to the case of classical wavelets, the transform coefficients in the quaternionic wavelet decomposition achieves \emph{energy compaction}. This means that (aside from the fact that the total energy in the original colour image would be equal to that of the decomposition) most of the energy in the quaternionic wavelet decomposition is concentrated in a few transform coefficients. More succinctly, let $F \in \mathbb{H}^{N\times N}$ be a colour image with $N\times N$ pixels whose RGB-NIR channels are embedded in the quaternion algebra. The \emph{energy $\xi_F$ of $F$} is given by $$\xi_F = \sum_{i,j=1}^N |F_{ij}|^2.$$ Furthermore, let $L^F_1 \geq L^F_2 \geq \cdots \geq L^F_{N^2}$ be the absolute value of the image pixels (treated as quaternions) of $F$ arranged in decreasing order. The \emph{cumulative energy profile of $F$} is given by $$\left(\frac{(L^F_1)^2}{\xi_F}, \frac{(L^F_1)^2 + (L^F_2)^2}{\xi_F}, \ldots, \frac{(L^F_1)^2 + (L^F_2)^2 + \cdots, (L^F_{N^2-1})^2}{\xi_F},1 \right).$$ The cumulative energy profile of the decomposition can be computed in a similar fashion.

\begin{figure}[h!]
	\centering
    \resizebox{0.5\textwidth}{!}{%
    \begin{tikzpicture}
    \begin{axis}[
    width=\linewidth,
        xmode=log,
        xmin=1,xmax=263000,
        ymin=0,ymax=1,
        scaled x ticks=false,
        x tick label style={/pgf/number format/fixed, /pgf/number format/set thousands separator={\,}},
        xminorticks=true,
        minor x tick num=1,
        yminorticks=true,
        log ticks with fixed point,
        minor y tick num=3,
        axis x line*=bottom,
        axis y line*=left,
        legend style={legend pos=south east,inner sep=0pt,outer sep=0pt,legend cell align=left,align=left,draw=none,fill=none,font=\normalsize}
        ]
    \addplot[color=violet, line width=2pt] table[x index=0, y index=1] {data.txt};
    \addlegendentry{Original image};
    \addplot[color=green, line width=2pt] table[x index=0, y index=2] {data.txt};
    \addlegendentry{Decomposition};
    \end{axis}
    \end{tikzpicture}}
	\caption{Cumulative energy profiles of the quaternion representation of the original RGB-NIR example image and its level 8 quaternionic wavelet decomposition. Notice that the energy in the wavelet decomposition is concentrated in a very few coefficients.}
	\label{fig:energyprofile}
\end{figure}
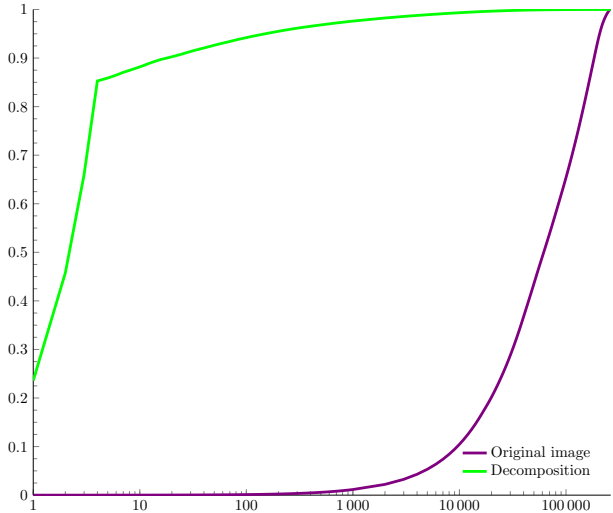

As an illustration, the cumulative energy profile of the sample RGB-NIR image and the cumulative energy profile of its level 8 wavelet decomposition are plotted and superimposed in Figure~\ref{fig:energyprofile}. Notice how the energy is compacted in only a very few transform coefficients in the decomposition. Through energy compaction, most of the signal's energy are concentrated in a small subset of significant wavelet coefficients, while less important details are represented by coefficients with lower magnitudes.

It is important to note that the level of energy compaction achieved through wavelet decomposition can vary based on the specific wavelet basis used, the nature of the signal, and the decomposition level. Some wavelet bases might provide better energy compaction for certain types of signals, while others might be more suitable for different applications.

Overall, energy compaction is a key reason why wavelet decomposition has found extensive use in signal and image processing tasks, offering efficient and effective ways to represent and manipulate data as we will see in the next section.

\section{Colour image processing with quaternionic wavelets}
\label{sec:applications}

The applicability of a wavelet transform in image processing is primarily rooted in its capacity to analyse images across varying scales while adeptly capturing both high and low-frequency components. By accommodating the multiresolution nature of images, wavelet transforms play a pivotal role in unveiling intricate patterns, detecting edges, mitigating noise, and preserving salient features. 

In this section, we delve into the basic yet profound capabilities of the quaternionic wavelet transform to elucidate its applicability in diverse image processing contexts and to highlight its role in extracting nuanced information from colour images. Our primary aim is to exhibit the potential of a holistic image processing methodology using quaternion-valued wavelets, with the specific focus on delineating their inherent promise rather than undertaking an assessment of their relative performance with respect to the conventional channel-by-channel approach.

\begin{table}[ht]
\centering
\resizebox{\textwidth}{!}{%
\begin{tabular}{|ll|c|c|c|c|c|}
\hline
\multicolumn{2}{|c|}{\textbf{Image Processing Step}}                  & \textbf{PSNR} & \textbf{SSIM} & \textbf{PIQE} & \textbf{BRISQUE} & \textbf{NIQE} \\ \hline
\multicolumn{1}{|l|}{\multirow{2}{*}{Compression}} & Original         & 221.42      & 1.00        & 38.50       & 8.32           & 2.19        \\ \cline{2-7} 
\multicolumn{1}{|l|}{}                             & Compressed       & 31.85       & 0.89        & 56.51       & 22.82          & 3.51        \\ \hline
\multicolumn{1}{|l|}{\multirow{2}{*}{Enhancement}} & Blurry           & 28.24       & 0.90        & 73.97       & 43.12          & 4.33        \\ \cline{2-7} 
\multicolumn{1}{|l|}{}                             & Enhanced         & 28.70       & 0.89        & 63.48       & 40.11          & 4.39        \\ \hline
\multicolumn{1}{|l|}{\multirow{3}{*}{Denoising}}   & Noisy            & 20.20       & 0.47        & 65.74       & 44.24          & 12.83       \\ \cline{2-7} 
\multicolumn{1}{|l|}{}                             & Soft-thresholded & 25.89       & 0.79        & 58.73       & 26.63          & 5.57        \\ \cline{2-7} 
\multicolumn{1}{|l|}{}                             & Hard-thresholded & 27.55       & 0.78        & 57.82       & 33.03          & 5.82        \\ \hline
\end{tabular}%
}
\caption{Summary of image quality scores for the original, blurred, noisy, and various reconstructed RGB-NIR images resulting from different image processing steps.}
\label{tab:image_scores}
\end{table}

\subsection{Compression}
The energy compaction in the quaternionic wavelet decomposition is a powerful property as it enables the extraction of the most informative aspects of a colour image while discarding less relevant information. This property aligns with the objectives of \emph{compression}: reducing data size, conserving storage space, and possibly optimising transmission bandwidth, all while maintaining the integrity and perceptual quality of the original content. 

In the image processing step of \cref{fig:QDWT}, we can perform a simple compression scheme by \emph{percentile thresholding} in the wavelet domain, i.e., zeroing out wavelet coefficients whose magnitudes are below a certain percentile. This results to  a sparse representation of the image in the wavelet domain while keeping the most important details preserved. Proceeding with the reconstruction step, yields a compressed version of the original image.

\begin{figure}[!h]
	\centering
	\centering
	\includegraphics[width=0.4\linewidth]{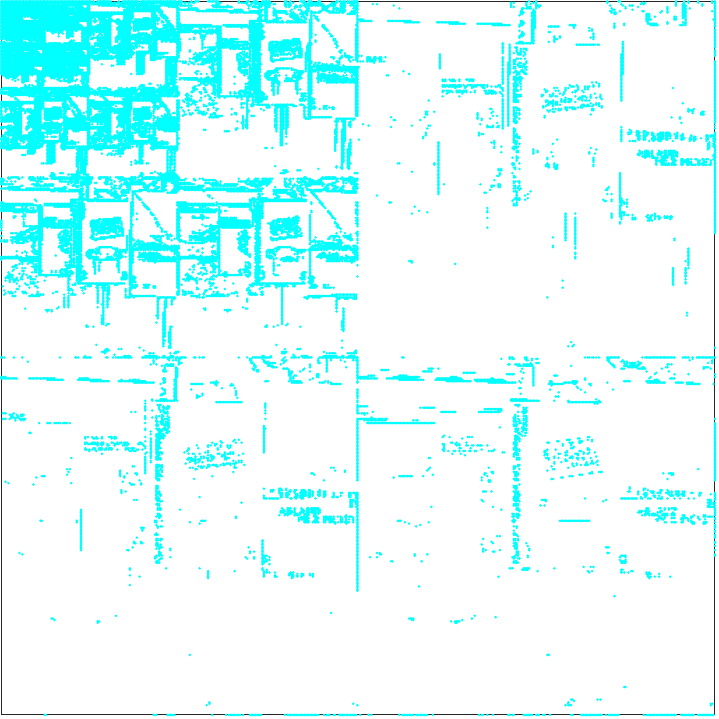}
	\caption{Location of the top 10\% wavelet coefficients that are kept after percentile thresholding in the quaternionic wavelet decomposition.}
	\label{fig:Nonzerocoefficients_quat}
\end{figure}

\begin{figure}[h!]
	\centering
    \includegraphics[width=0.8\linewidth]{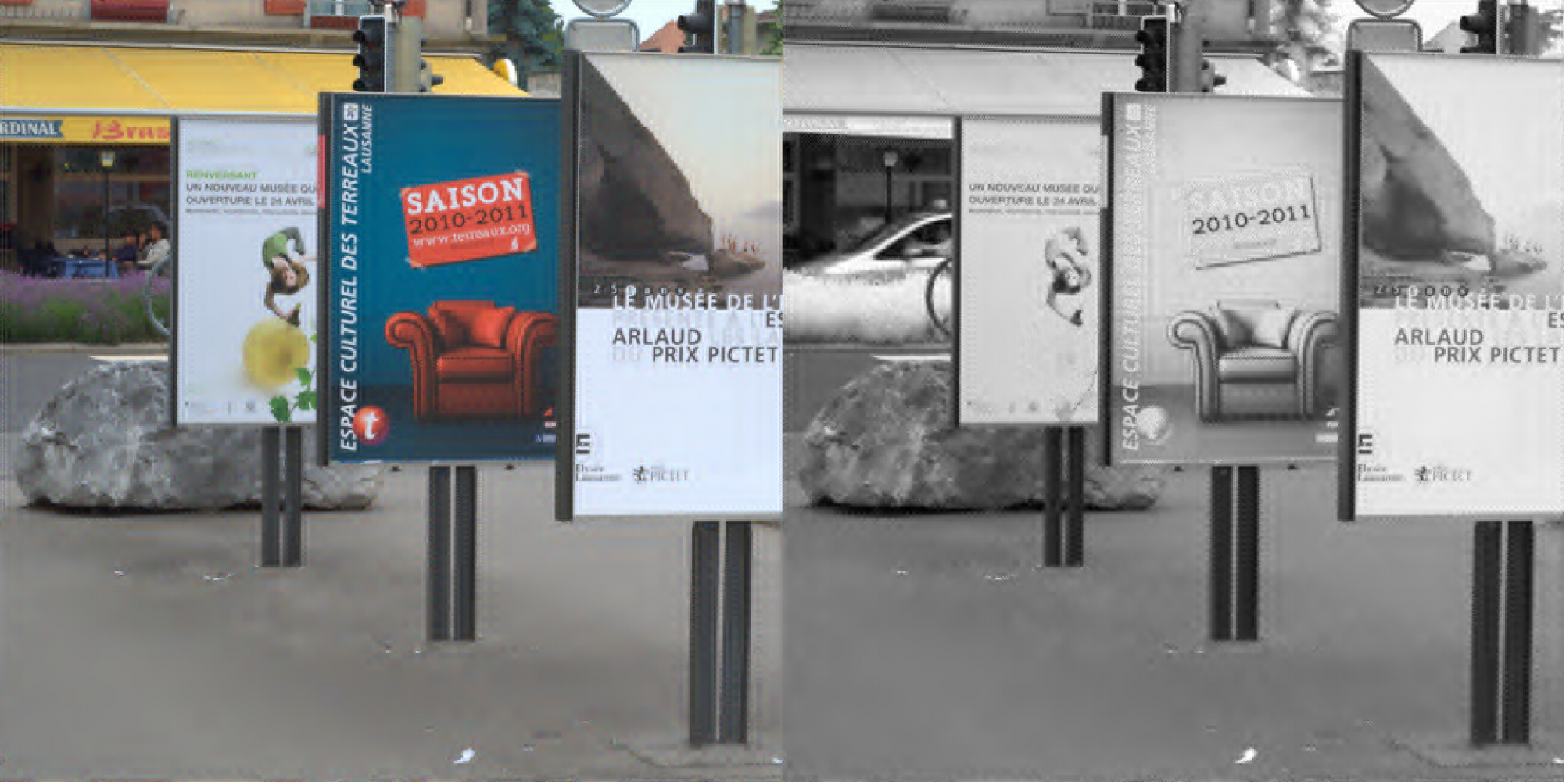}\hfill
	\caption{RGB-NIR image reconstructed from a level 8 quaternionic wavelet decomposition, thresholded to use only the top 10\% of the wavelet coefficients.}
	\label{fig:Quat_compressed}
\end{figure}

Exemplifying this with our given RGB-NIR image, we apply a threshold to its level 8 wavelet decomposition to keep only the top 10\% of the wavelet coefficients. \cref{fig:Nonzerocoefficients_quat} shows the location of the wavelet coefficients that were retained after applying the threshold. The remaining wavelet coefficients are then reconstructed to obtain a compressed image given in \cref{fig:Quat_compressed}. The image quality scores of the compressed RGB-NIR image are given in \cref{tab:image_scores}. Comparing these scores with that of the original RGB-NIR image, it is evident that the compressed RGB-NIR image maintains favourable scores in no-reference quality metrics even under substantial compression.

\begin{figure}[h!]
	\centering
	\centering
	\includegraphics[width=0.8\linewidth]{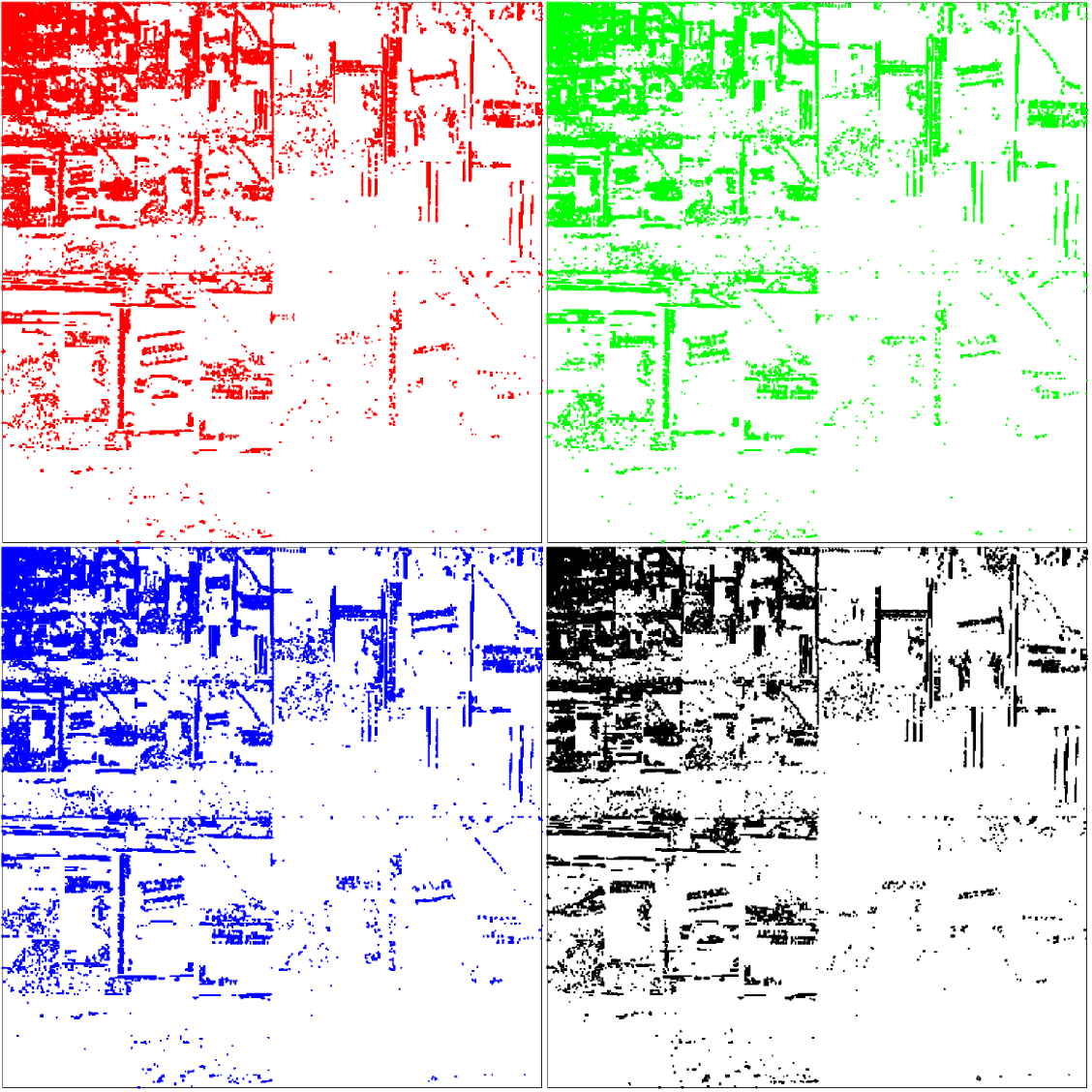}
	\caption{Location of the top 10\% wavelet coefficients (in the red, green, blue and near-infrared channels) that are kept after percentile thresholding in the channel-by-channel wavelet decomposition. The points that correspond to the remaining nonzero wavelet coefficients in each channel are evidently present at different locations.}
	\label{fig:channelthresholding}
\end{figure}

Recall that tensored Daubechies' wavelets may also be used to perform wavelet decomposition of the RGB-NIR channels separately. This process produces four wavelet decompositions --- one for each channel. We may also perform a suitable amount of percentile thresholding on these four sets wavelet coefficients separately. After such, a reconstruction on each channel is performed to result to a compressed version of the original image. While energy compaction also happens in the wavelet decomposition of each channel, the retained coefficients after  thresholding are at different locations for each channel (see \cref{fig:channelthresholding}).

In view of \cref{fig:Nonzerocoefficients_quat}, we note that the count of thresholded quaternionic wavelet coefficients is only at $262,144$ locations. In contrast, looking at \cref{fig:channelthresholding} where we used channel-by-channel decomposition and thresholding, the total number of locations for real wavelet coefficients is four times that of the quaternionic approach. This makes keeping track of the location of nonzero wavelet coefficients more expensive in the channel-by-channel decomposition as compared to when the quaternionic wavelet decomposition and thresholding is done.  Consequently, enhanced compression becomes attainable through the use of quaternion-valued wavelets as the spatial distribution of thresholded coefficients is rendered consistent across all channels. This alignment is anticipated to yield memory conservation within the position encoding (of wavelet coefficients) step of a conventional wavelet-based compression framework.

It is also worth mentioning that the effect of non-separability turned up in the thresholded quaternionic wavelet decomposition. In \cref{fig:Nonzerocoefficients_quat}, there is a fair distribution of nonzero values in the detail coefficients. On the other hand, when employing tensored Daubechies' wavelets, the thresholded wavelet coefficients exhibit fewer non-zero values for the diagonal details, as evident in \cref{fig:channelthresholding}.



\subsection{Image enhancement}

Wavelet transforms can be used to enhance certain features of an image. By modifying the wavelet coefficients, we can amplify or suppress specific frequency components to improve visual quality or emphasise certain image characteristics.

In the image processing step of \cref{fig:QDWT}, we carry out image enhancement by simply multiplying the detail coefficients by a constant greater than one while leaving the approximation coefficients unchanged. Reconstructing from these updated coefficients produces an image with enhanced edges. More pronounced edges can be obtained by accentuating the detail coefficients using larger multipliers.

\begin{figure}[hbt!]
	\centering
    \includegraphics[width=0.8\linewidth]{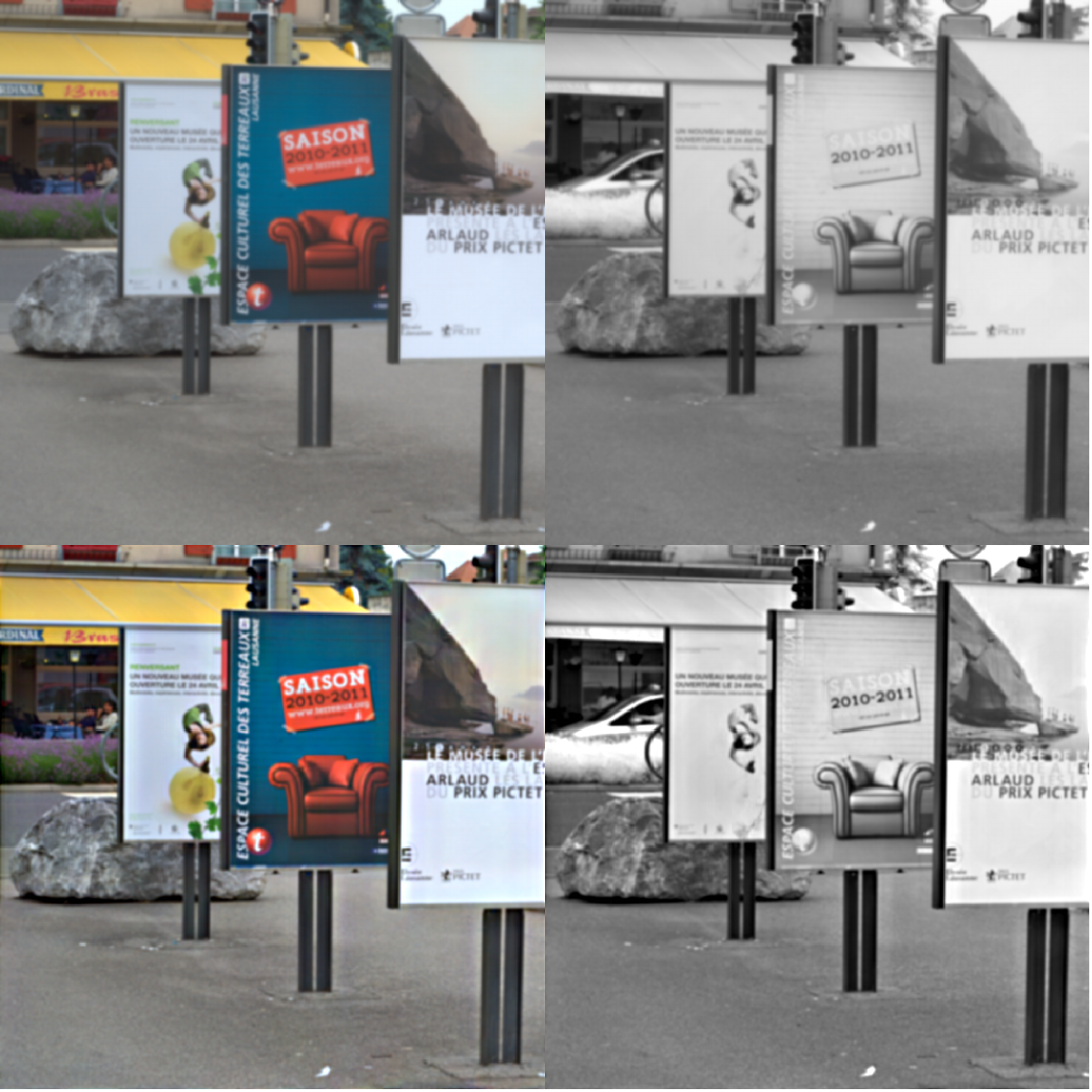}\hfill
	\caption{Blurred RGB-NIR image (top) and the enhanced RGB-NIR image (bottom) using quaternionic wavelets.}
	\label{fig:enhanced}
\end{figure}

For an example, refer to \cref{fig:enhanced} (top RGB-NIR image pair). Here, we started with our standard RGB-NIR image example but blurred it with a Gaussian filter with a standard deviation of 1.25. From its level 8 wavelet decomposition, we amplify the detail coefficients  multiplying it by a factor of 1.25, while leaving the approximation coefficients the same. Proceeding with the reconstruction step yields the sharpened image in \cref{fig:enhanced} (bottom RGB-NIR image pair).  We see in this simple illustration that wavelet-based enhancement techniques can help in sharpening the edges and adjusting the contrast of a blurred image. By enhancing the high-frequency components, edges and boundaries become more distinct, also leading to an overall improvement in image contrast. The image quality scores of the blurred and the enhanced RGB-NIR colour images are given in \cref{tab:image_scores}.

Wavelet-based image enhancement finds applications in diverse domains. In medical imaging, it can aid in diagnosing diseases by making subtle details in scans more evident. In satellite imagery, it can unveil hidden geographical features. In art restoration, it can enhance aged or deteriorated images.

\subsection{Edge detection}

The wavelet transform, with its multi-scale decomposition, can provide a robust approach to edge detection. During wavelet decomposition, the high-frequency information associated with edges is captured within the detail coefficients. As the decomposition progresses to higher scales, these coefficients represent increasingly fine variations in the image. Thus, the detail coefficients highlight the high-frequency components, effectively pinpointing edges within the image. This idea presents a simple edge detection scheme. 

In the image processing step of \cref{fig:QDWT}, we perform an edge detection scheme by completely discarding the approximation coefficients while retaining the detail coefficients. Reconstructing from the remaining wavelet coefficients yields the edges in the originally given colour image. 

\begin{figure}[hbt!]
	\centering
    \includegraphics[width=0.8\linewidth]{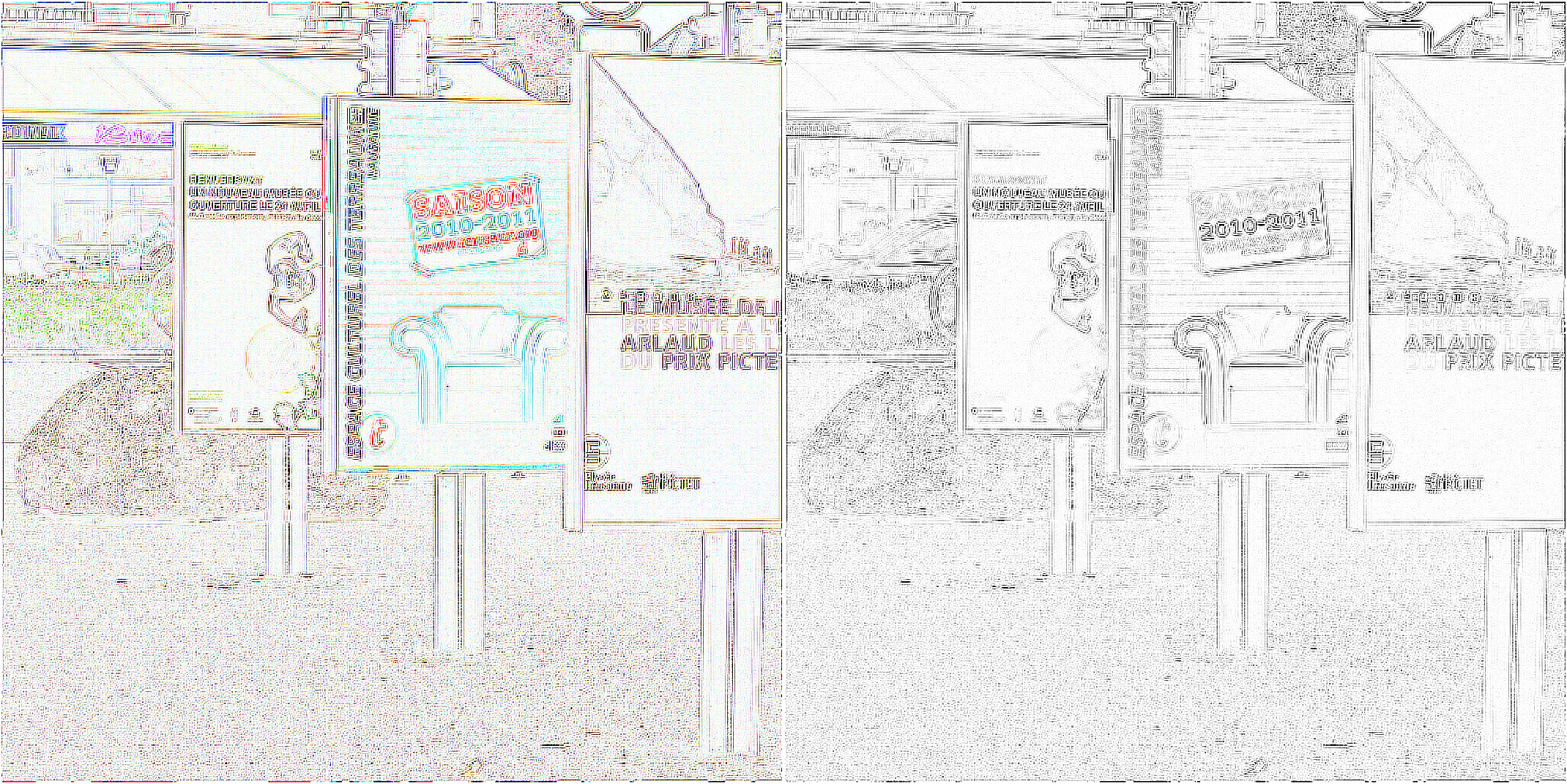}\hfill
	\caption{Edges detected by the quaternionic wavelets in the RGB (left) and NIR (right) components of the sample RGB-NIR image.}
	\label{fig:edges}
\end{figure}

An example that uses this basic edge detection scheme is illustrated in \cref{fig:edges}. The resulting edges are worked out from a level 6 quaternionic wavelet decomposition of our given RGB-NIR image. While straightforward, this method is evidently not yet optimal since the reconstruction still includes unnecessary details. Further improvement can be achieved by applying advanced thresholding techniques to the detail coefficients to better capture significant details before performing the reconstruction step.

When applying wavelet-based edge detection, it is important to consider the choice of wavelets, the decomposition level, and the thresholding or enhancement techniques. These choices influence the accuracy of edge detection and the quality of the image results. Furthermore, integrating wavelet-based edge detection with other image processing methods can result in a more thorough and proficient extraction of edges within intricate images.

In the literature, edge detection methods that use shearlets and curvelets are recognised for their superior performance compared to wavelet-based approaches \cite{shearlet,curvelet}. This is attributed to the effectiveness of shearlets and curvelets in accurately representing images with edges, as they capture multidirectional features. Nevertheless, wavelet-based edge detection finds applications in numerous fields, including computer vision, medical imaging, and remote sensing. In medical imaging, accurate edge detection assists in segmenting organs or structures, aiding in diagnosis. In object recognition, it helps identify shapes and patterns, forming the basis for more complex analysis and feature extraction tasks.

\subsection{Denoising}

Since wavelets can provide a multiresolution representation of a signal, they are able to capture both fine-scale and coarse-scale details of the signal. Such is crucial for denoising because noise often affects different scales of a signal differently. This property, combined with effective thresholding strategies, enables wavelets to separate noise from the true image features.

In the image processing step of \cref{fig:QDWT}, we denoise a noisy colour image by applying a threshold to the wavelet coefficients to remove noise while keeping important image details. This thresholding step is a crucial step to separate noise from the true image features. There are different methods for thresholding, and each of these has their own approach to determining which coefficients to keep and to discard. After the thresholding step, we perform a reconstruction from the remaining wavelet coefficients to yield a denoised version of the noisy colour image. The quality of denoising using wavelet decomposition depends on various factors including the choice of wavelets, thresholding method, threshold values, and the characteristics of the noise in the image.


The thresholding step is often carried out by using either a \emph{universal} threshold or an \emph{adaptive} threshold. Universal thresholding involves applying the same threshold value to all the coefficients in a particular wavelet subband while adaptive thresholding sets different threshold values for different subbands based on their characteristics. The latter takes into account that different subbands might contain varying amounts of noise and signal information.

\begin{figure}[hbt!]
	\centering
    \includegraphics[width=\linewidth]{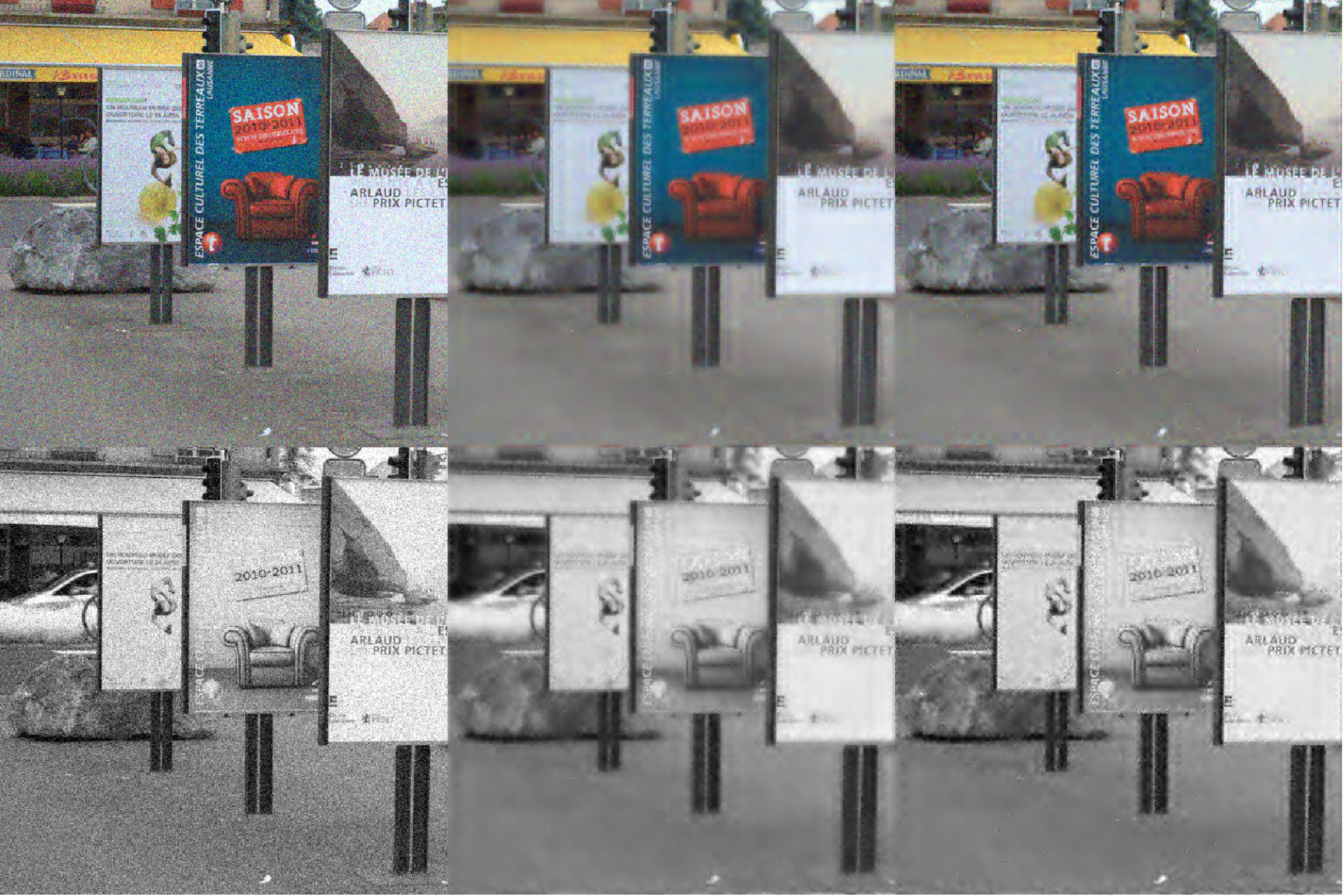}\hfill
	\caption{First row (from left to right): noisy RGB, soft-thresholding denoised RGB part, hard-thresholding denoised RGB part. Second row (from left to right): noisy NIR, soft-thresholding denoised NIR part, hard-thresholding denoised NIR part.}
	\label{fig:softthresh}
\end{figure}

For a simple illustration on how quaternion-valued wavelets can be applied to denoising, we only exemplify with universal thresholding. For instance, we consider \emph{VisuShrink} which follows a universal threshold $t = \sigma \sqrt{2 \log n}$ where $\sigma^2$ is the noise variance and $n$ is the number of pixels \cite{donoho}. This universal threshold  may be used to perform either a \emph{soft} or \emph{hard} thresholding which we describe through the \emph{soft-thresholding} $S_t: \mathbb{H} \to \mathbb{H} $ and \emph{hard-thresholding} $H_t: \mathbb{H}  \to \mathbb{H} $ operators defined by 
\[
S_t(x) = \begin{cases}
        \dfrac{x}{|x|}\max(|x|-t,0) & \quad |x| \neq 0\\  0 &  \quad |x|=0 \end{cases}  \qquad \text{and} \qquad H_t(x) = \begin{cases}
		x & \quad |x|>t \\
		0 & \quad \text{otherwise}
\end{cases},\]
respectively. Note that the definition of $S_t$ and $H_t$ are modified from their classical definitions to be able to handle quaternion values. For an example, refer to \cref{fig:softthresh}. The noisy RGB-NIR is corrupted by Gaussian noise with standard deviation of 0.1. We used both thresholding techniques on a level 4 quaternionic wavelet decomposition. The image quality scores of the noisy, soft-thresholding denoised and hard-thresholding denoised RGB-NIR images are summarised in \cref{tab:image_scores}. The denoised RGB-NIR image showed significantly enhanced image quality scores from that of the noisy colour image. However, it is important to observe that the application of soft-thresholding carries the risk of smoothing out crucial features, potentially leading to a denoised image that appears blurry. On the other hand, the hard-thresholding approach may exhibit excessive aggressiveness, running the risk of discarding coefficients integral to the signal -- such aggressiveness can cause distortion in the reconstructed signal.

Our preliminary investigations revealed that better denoised images are obtainable with adaptive thresholding schemes. However, it is not yet clear how to choose these threshold values to declare optimal results. Such superiority of adaptive thresholding is somehow expected as these methods take into account the inherent variability of the signal and noise and adjust the threshold accordingly.

\section{Conclusion}

The successful construction of compactly supported, smooth, and orthonormal quaternion-valued wavelets on the plane has paved the way for numerous critical avenues of exploration. With these wavelets, the opportunity arises to encode the constituent elements of a colour image within the scalar and imaginary parts of quaternions, enabling comprehensive signal processing through wavelet transforms. Quaternion-valued wavelets on the plane hold great promise as a transformative tool in colour image processing, offering a holistic wavelet-based approach that harnesses inter-channel relationships for more accurate and efficient image analysis and enhancement. 

The proposed scheme for colour image decomposition and reconstruction using quaternionic scaling and wavelet filters demonstrated perfect reconstruction and efficiency of energy compaction. Additionally, the exemplified image processing steps for compression, denoising, segmentation, and enhancement underscored the versatility of quaternion-valued wavelets in addressing a spectrum of colour image processing applications. In particular, better compression is viable with quaternion-valued wavelets since the location of thresholded coefficients are no longer different from each channel. This is expected to save memory in the position encoding (of wavelet coefficients) step of a wavelet-based compression scheme. Determining whether this holistic approach to colour image enhancement, edge detection, and denoising performs better than a channel-by-channel approach remains a promising direction for investigation. It is worth looking into optimising the execution of these basic image processing steps that we have illustrated. For instance, level-dependent thresholding can be investigated for edge detection, denoising and enhancement.  Once optimised, the effectiveness of such approaches can be compared to representative state-of-the-art image processing methods at disposal. Additionally, it is interesting to explore the potential of quaternion-valued wavelets in sparsity-promoting wavelet-based regularisation for denoising and deblurring, and in general inverse problems. However, variational formulation of inverse problems using quaternions would also require the development of end-to-end quaternion optimisation frameworks, e.g., quaternion alternating direction of multipliers method \cite{miron2023quat,qADMM}. Furthermore, given the emergence of quaternion neural networks, it would be interesting to examine the potential applicability of these quaternion-valued wavelets in such a landscape \cite{miron2023quat,cnnwavelets}. In general, the inherent simplicity of applying quaternion-valued wavelets to foundational image processing tasks calls for further exploration in advanced image processing, mirroring the influence that  classical real-valued wavelets have had.

\section*{Acknowledgements}

NDD and JAH were supported by Australian Research Council Grant DP160101537. NDD was supported in part by an AustMS Lift-Off Fellowship.


\section*{About the authors}

\noindent \textbf{Neil D. Dizon} (neil.dizon@helsinki.fi) received his Ph.D. in Mathematics from the University of Newcastle in Australia. He is currently a Postdoctoral Researcher at the University of Helsinki in Finland. His research interests include wavelet analysis and continuous optimisation.\\

\noindent \textbf{Jeffrey A. Hogan} (jeff.hogan@newcastle.edu.au) received his Ph.D. in Mathematics from the University of New South Wales. He is currently an Associate Professor at the University of Newcastle in New South Wales, Australia. His research interests include Clifford--Fourier theory, prolate spheroidal wave functions, and wavelets.

\end{document}